\documentclass{article}
\pdfoutput=1

\usepackage[preprint]{neurips}
\usepackage[utf8]{inputenc} % allow utf-8 input
\usepackage[T1]{fontenc}    % use 8-bit T1 fonts
\usepackage{hyperref}       % hyperlinks
\usepackage{url}            % simple URL typesetting
\usepackage{booktabs}       % professional-quality tables
\usepackage{amsfonts}       % blackboard math symbols
\usepackage{microtype}      % microtypography
\usepackage{xcolor}         % colors
\usepackage{natbib}
\usepackage{graphicx}
\usepackage{algorithm}
\usepackage{algorithmic}
\usepackage[marginal]{footmisc}

\setcitestyle{numbers,square}

\title{Efficient and Accurate Co-Visible Region Localization with Matching Key-Points Crop (MKPC): A Two-Stage Pipeline for Enhancing Image Matching Performance}

\author{%
	Hongjian Song*, Yuki Kashiwaba*, Shuai Wu, Canming Wang}
\begin{document}
	
	\maketitle

	\begin{abstract}
		Image matching is a classic and fundamental task in computer vision. In this paper, under the hypothesis that the areas outside the co-visible regions carry little information, we propose a matching key-points crop (MKPC) algorithm. The MKPC locates, proposes and crops the critical regions, which are the co-visible areas with great efficiency and accuracy. Furthermore, building upon MKPC, we propose a general two-stage pipeline for image matching, which is compatible to any image matching models or combinations. We experimented with plugging SuperPoint + SuperGlue into the two-stage pipeline, whose results show that our method enhances the performance for outdoor pose estimations. What's more, in a fair comparative condition, our method outperforms the SOTA on Image Matching Challenge 2022 Benchmark, which represents the hardest outdoor benchmark of image matching currently.\footnote{* denotes contributing equally to this work.\\}
	\end{abstract}
	
	\section{Introduction}
	
	Image matching is a fundamental and classic task in computer vision, playing a crucial role in various downstream tasks such as Simultaneous Localization and Mapping (SLAM), Structure-from-Motion (SfM), and Localization. In the task of image matching, the objective is to establish the relationship between two given images by employing methods such as feature point extraction~\cite{detone2018superpoint, tyszkiewicz2020disk}, feature point description~\cite{detone2018superpoint, tyszkiewicz2020disk}, and feature point matching~\cite{sarlin2020superglue}. This relationship can be described using a fundamental matrix, an essential matrix, or a homography matrix. Recently, deep learning models have been developed to implement end-to-end keypoint matching, capable of outputting matched key-points by simply inputting two images~\cite{edstedt2022deep, sun2021loftr, tang2022quadtree, chen2022aspanformer}.

\begin{figure*}[!h]
	\centering
	%\fbox{\rule[-.5cm]{0cm}{4cm}  \rule[-.5cm]{4cm}{0cm}}
	\includegraphics[scale=0.2]{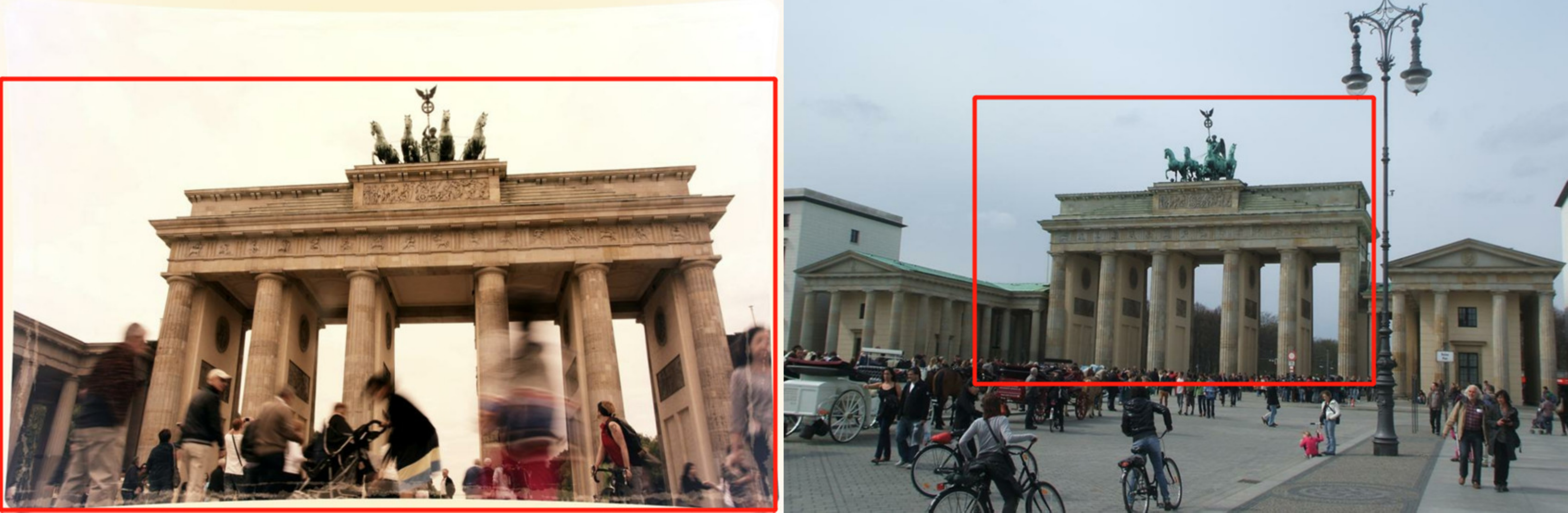}
	\caption{An example of critical regions of two images}
	\label{demo}
	\end{figure*}

	While most current works focus on interest point detectors~\cite{detone2018superpoint, tyszkiewicz2020disk}, descriptors~\cite{detone2018superpoint, tyszkiewicz2020disk}, matching, or implementing in an end-to-end way~\cite{edstedt2022deep, sun2021loftr, tang2022quadtree, chen2022aspanformer}. we found that the input images have a significant impact on the performance of image matching. Essentially, improving the performance of image matching primarily involves reducing the number of incorrect matches while increasing the number of correct matches. Under the assumption that areas outside the co-visible regions contain little valuable information and the matching points outside these regions are incorrect matches, we define the co-visible area of two images as the "critical region" (Figure \ref{demo}). By focusing on the critical region as the matching target, we aim to enhance image matching performance. Identifying the critical regions shared between two images and applying fine-grained matching within these regions offers two significant advantages: (i) it refines the matching of critical regions and (ii) it rejects the majority of outliers or noisy points. Based on these considerations, we propose a simple but effective algorithm in image matching, called Matching Key-Points Crop (MKPC). The essence of MKPC is a DBSCAN clustering algorithm. The MKPC algorithm leverages matching key-points obtained from arbitrary matching models to efficiently and accurately crop the critical regions of both images with great efficiency and remarkable accuracy, while the key-points produced by upstream models are representative. 

	Building upon MKPC, we propose a two-stage pipeline for image matching that can be easily extended for any image matching models while preserving superior performance. In the first stage, arbitrary image matching models generate matching points for the two images. Subsequently, the MKPC algorithm clusters these matching points and proposes a Region of Interest (ROI) area, which represents the critical regions for both images. A naive array slicing operation crops the critical regions, which are then fed into arbitrary image matching models after necessary pre-processing. This constitutes the second stage of our two-stage pipeline. Our experiments demonstrate that the two-stage pipeline with MKPC as the core consistently improves image matching performance on the PhotoTourism dataset, which is a subset of YFCC100M dataset~\cite{thomee2016yfcc100m} with ground truth poses. The experimental results not only validate the effectiveness of our proposed method, including MKPC and the two-stage pipeline but also support the theoretical foundation that the key to enhancing image matching performance lies in accurately locating the co-visible areas. Although our method is also applicable to indoor scenarios, it requires higher computational complexity. Furthermore, our approach outperforms the state-of-the-art (SOTA) on the Image Matching Challenge 2022 benchmark, which is considered the most challenging outdoor pose estimation dataset.
	
	The contributions of this paper are as follows: (i) We propose that the key to enhancing the performance of image matching lies in identifying critical regions, or the co-visible areas in both images. (ii) We introduce Matching Key-points Crop(MKPC), a critical region finder that efficiently crops these areas with remarkable accuracy, and (iii) Building upon MKPC, we present a two-stage pipeline that enhances the effectiveness of arbitrary image matching models for outdoor pose estimations. In addition, our proposed two-stage pipeline achieved a higher score than the SOTA score on the Image Matching Challenge 2022 benchmark, which represents the most difficult benchmark for outdoor estimation currently.

	\section{Method}
	
	Defining the co-visible area between two images as the critical region, we propose that the key to improve the effect of image matching is to find the critical region with better accuracy. We propose this point based on the assumption that the co-visible area does not contain information useful for image matching outside that region. Therefore, we propose a matching key-points crop(MKPC) algorithm, which is capable to locate the critical region effectively and accurately. Building upon MKPC, we further propose a two-stage pipeline which supports being extended arbitrarily.
	
	\subsection{Matching Key-points Crop (MKPC)}
\begin{figure*}[!h]
	\centering
	%\fbox{\rule[-.5cm]{0cm}{4cm}  \rule[-.5cm]{4cm}{0cm}}
	\includegraphics[scale=0.34]{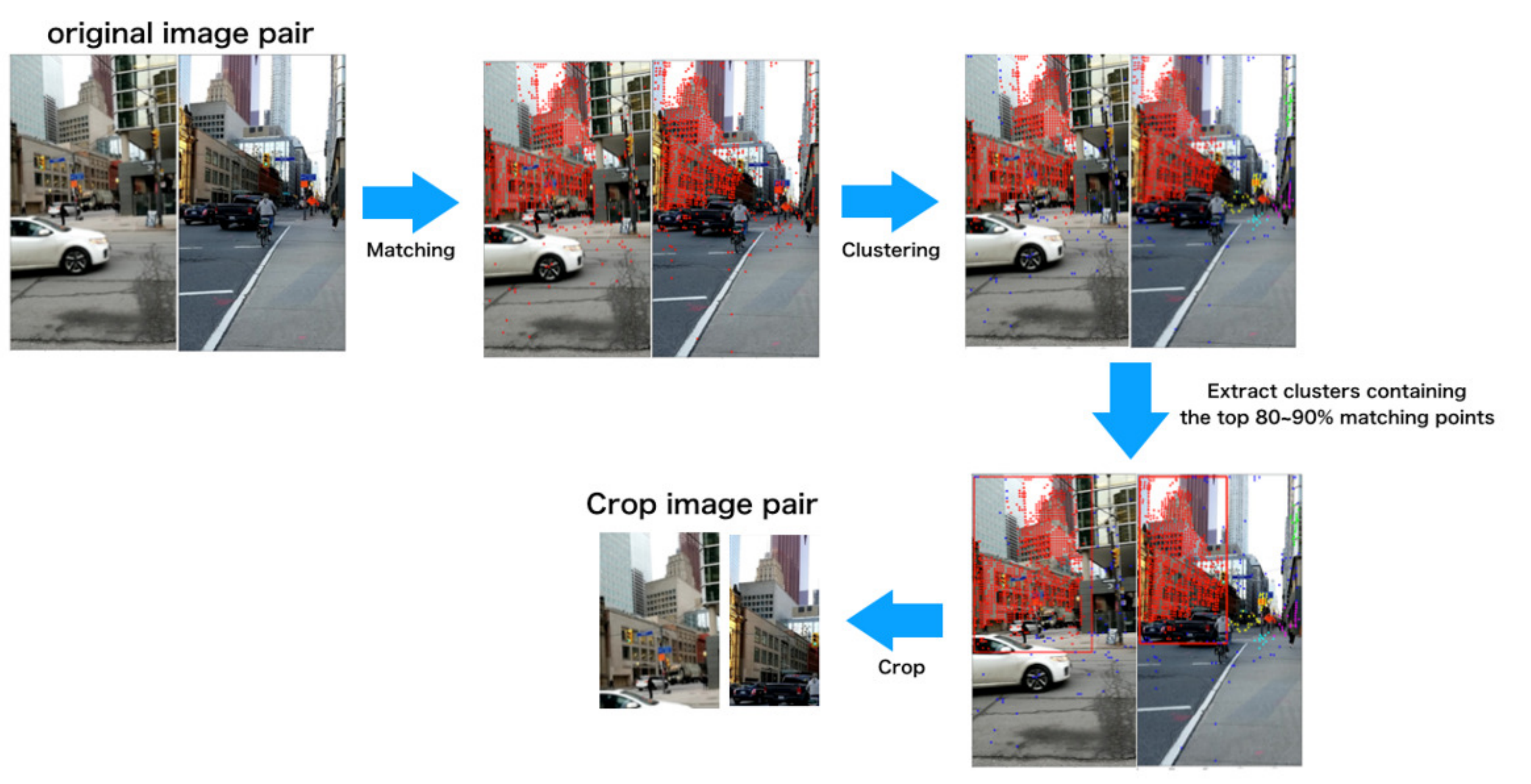}
	\caption{The workflow of matching key-points crop(MKPC)}
	\label{mkpc}
	\end{figure*}

	The MKPC algorithm crops critical regions by clustering the matching key-points between two images outputted by arbitrary image matching models. The workflow is shown is Figure \ref{mkpc}. Before applying MKPC, we already have two images $I_1$ and $I_2$, and matched key-points on both images with any model (or any combinations of models), which is defined as $X^1_1$ and $X^1_2$, respectively. $X^1_i$ denotes the $i_{th}$ image in the first stage of the two-stage pipeline which will be proposed in the next subsection. It consists of the following three steps: (i) Clustering the matching key-points($X^1_1$ and $X^1_2$) of two images($I_1$ and $I_2$) with DBSCAN. (ii) Generate a bounding box by filtering and gathering those clusters (iii) Crop the area covered by the Bounding box. The Algorithm \ref{alg1} describes the MKPC algorithm flow in detail. With the input of two images ($I_1$, $I_2$)  and corresponding matched key-points in stage-one ($X^1_1$ and $X^1_2$), the MKPC outputs the respective cropped critical regions.
	
	We choose DBSCAN, instead of other cluster algorithms such as K-means, to implement MKPC for some of its structural advantages. (i) DBSCAN is a density-based method, which intuitively suitable for this work. That is, areas with high-density matching key-points must be important, while places with low density may be not so important or even outliers. (ii) DBSCAN is robust to outliers, and the one of the motivations of DBSCAN is rejecting outliers. (iii) we never needs to specify the number of clusters before applying a DBSCAN, while we needs before applying other clusters such as K-means. 
	
	\begin{algorithm}
		\caption{The Workflow of Matching key-points Crop (MKPC)}
		\label{alg1}
		\begin{algorithmic}
			\REQUIRE ~~\\
			Two images to be matched $I_1$ and $I_2$; \\
			The matched key-points on both images with any models (or combinations of models) $X^1_1$ and $X^1_2$; \\
			A cluster selection threshold $T$, which is set to $0.05$ in our experiment;\\
	        \ENSURE ~~\\
	        (1) Clustering on $X^1_1$ and $X^1_2$ with a DBSCAN to get collections of clusters $C_1$ and $C_2$ for both images; \\
	        (2) Sorted the clusters in $C_1$ and $C_2$ in descending order according to the number of points in each cluster. Supposing clusters $C_1$ and $C_2$ contained $k1$ and $k2$ clusters, respectively, we define $C_1=\{c^1_1, c^2_1, \cdots, c^{k1}_1\}$ and $C_2=\{c^1_2, c^2_2, \cdots, c^{k2}_2\}$, in which $c^1_1$ and $c^1_2$ denotes the "largest" cluster in $C_1$ and $C_2$, respectively; \\
	        (3) Define two empty candidate clustering sets $C'_1$ and $C'_2$ for each of the two images; \\
	        (4) Select the largest size cluster $c^1_1$ for $C_1$. Append $c^1_1$ into $C'_1$; \\
	        (5) Define the current selected index as $i=2$, Select the next largest cluster $c_{next}=c^i_1$; \\
	        	 \quad \ \  Get the ratio $r$ between $c_{next}$ and $c^1_1$, $r_1=\frac{c_{next}}{c^1_1}$; \\
	        	 \quad \ \  while ($r>=T$ and $i<=k1$) \ do \\
	        	 \quad \ \  \quad \ \ Append $c^{next}$ into $C'_1$; \\
	        	 \quad \ \  \quad \ \ $i=i+1$; \\
	        	 \quad \ \  \quad \ \ Select the next largest cluster $c_{next}=c^i_1$; \\ 
	        	 \quad \ \  \quad \ \ Get the ratio $r$ between $c_{next}$ and $c^1_1$, $r_1=\frac{c_{next}}{c^1_1}$; \\
	        (6) Apply the same operations as step (4) and step (5) for another image $I_2$ and obtain the candidate clusters set $C'_2$; \\
	        (7) Take the 4 points $P_1$ in the plane coordinate system that are the most upper-left, the most lower-left, the most upper-right, and the most lower-right of all points in the all the clusters in $C'_1$. Apply the same operations for $C'_2$ to get $P_2$;\\
	        (8) Use $P_1$ to crop $I_1$ to get $I'_1$. Use $P_1$ to crop $I_2$ to get $I'_2$; \\
	        (9) Return $I'_1$ and $I'_2$; \\

		\end{algorithmic}
	\end{algorithm}

\begin{figure*}[!h]
	\centering
	%\fbox{\rule[-.5cm]{0cm}{4cm}  \rule[-.5cm]{4cm}{0cm}}
	\includegraphics[scale=0.2]{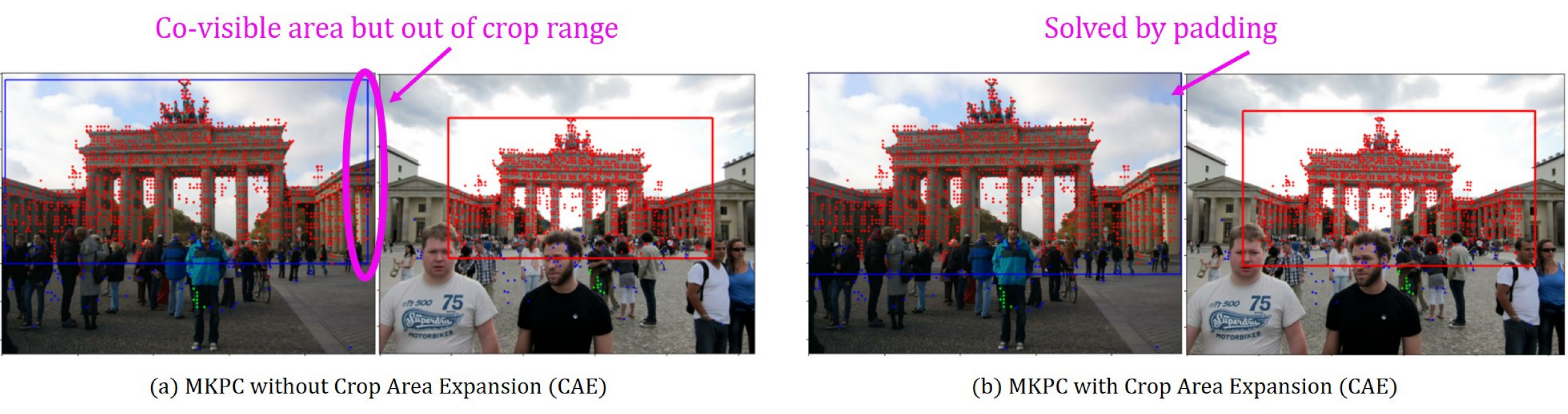}
	\caption{A demo of matching key-points crop(MKPC).After the loop in \ref{alg1}, the green, yellow, and red cluster will be gathered, while the pink one and the blue one will be rejected. If the pink one is included, more areas without useful information will be included, making rejected the pink cluster makes sense.}
	\label{mkpc_demo}
	\end{figure*}

	Take Figure \ref{mkpc_demo} as an example. After clustering, $X^1_1$ is split into four clusters (red with 235 points, green with 894 points, yellow with 471 points, and pink with 9 points) and outliers (blue), and $X^1_2$ is split into two clusters(red, green) and outliers (blue). For $I_2$, it's obvious that the red and green clusters will be gather and the outliers will be rejected. For $I_1$, in the first loop, MKPC selects the green one with the largest clusters and the yellow clusters as the second largest cluster, and both are extended into the candidate clustering sets $C'_1$ after the judgment. Afterwards, the red one is extended while the pink one is rejected. Hence, after the loop in Algorithm \ref{alg1}, the green, yellow, and red clusters are gathered, while the pink one and the blue one are rejected.
	
	\subsection{Two-stage Pipeline of Image Matching}
	
\begin{figure*}[!h]
		\centering
		%\fbox{\rule[-.5cm]{0cm}{4cm}  \rule[-.5cm]{4cm}{0cm}}
		\includegraphics[scale=0.2]{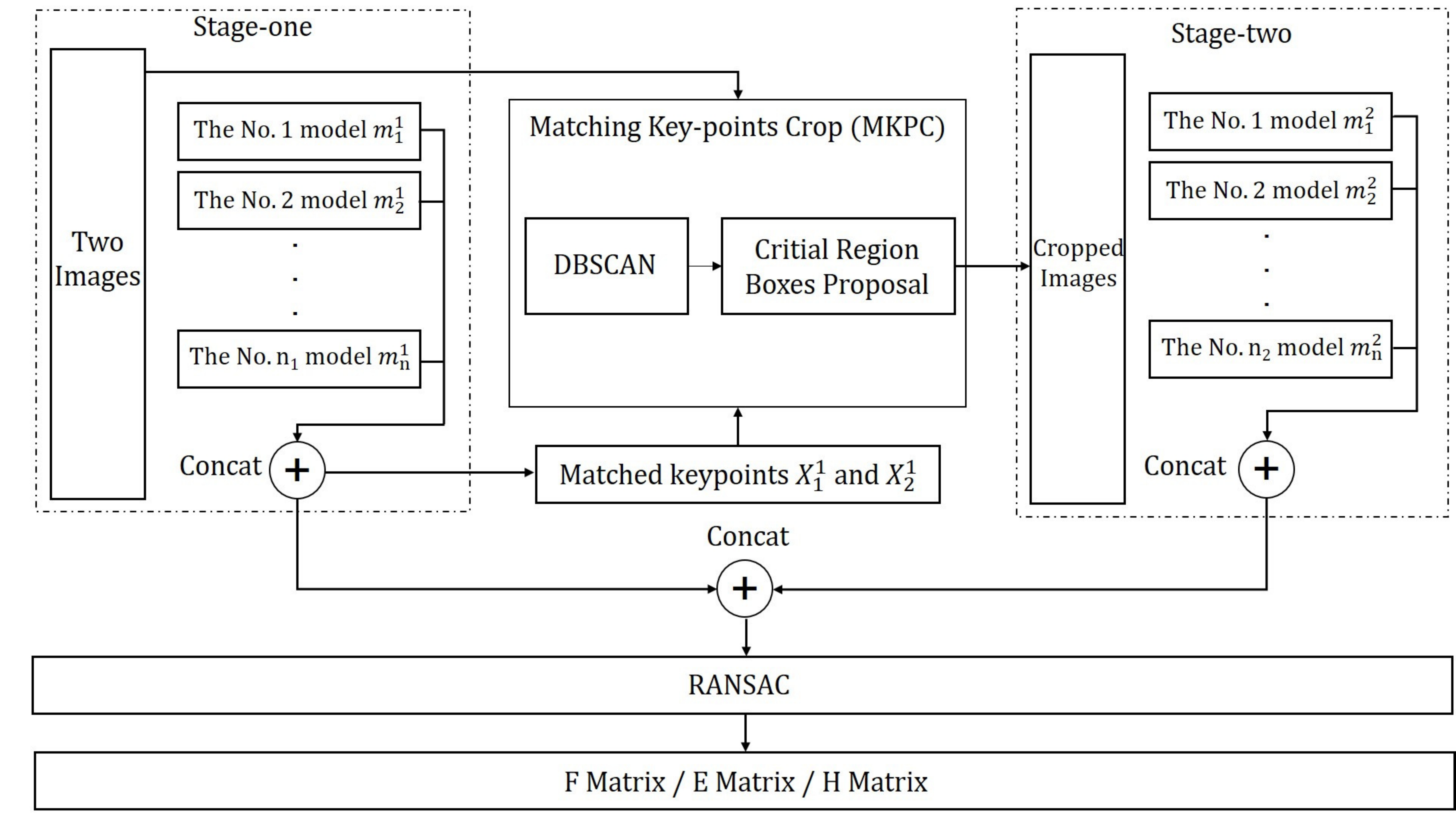}
		\caption{The workflow of matching key-points crop(MKPC)}
		\label{2stage}
	\end{figure*}

	With MKPC as the core, we propose an arbitrarily extensible two-stage pipeline for image matching. As shown in Figure \ref{2stage}, The two-stages pipeline support plugging any image matching models or recipe of models into both stages, such that the models set for stage 1 is $M^1=\{m^1_1, m^1_2, \cdots, m^1_{n_1} \}$, and the models set for stage 2 is $M^2=\{m^2_1, m^2_2, \cdots, m^2_{n_2} \}$. Each of the models in stage-one produce a set of key-points for both images. The concatenated key-points of stage-one $X_1^1$ and $X_2^1$ are then fed into MKPC algorithm. Based on the two images $I_1$ and $I_2$ and the key-points of stage-one $X_1^1$ and $X_2^1$, The MKPC generates the cropped critical regions for both images ($I'_1$ and $I'_2$). With the cropped regions of both images, each of the models in stage-two outputs the matched key-points for stage-two ($X_1^2$ and $X_2^2$). Afterwards, the models in stage-two matches on the cropped area, producing the key-points for the second stage $X^2_1$ and $X^2_2$. Concatenating the key-points of both stages, i.e. getting the matched key-points for the first image $X_1 = Cat(X^1_1, X^2_1)$ and the matched key-points for the second image $X_2 = Cat(X^1_2, X^2_2)$, two sets of matching key-points for both images are generated. Both of the sets are then fed into a RANSAC algorithm to maintain the F matrix, E matrix, or H matrix, as illustrated in Algorithm \ref{alg2}. Including the key-points of stage-one may seem contradictory to our assumption that little valuable information is outside the critical regions. However, MKPC suffers a risk of removing small co-visible regions because it discards clusters with few matching points. Although including the key-points in stage-1 potentially absorbs outliers, the advantage of encompassing the matching points that stage-2 may miss is greater. The percentage of correct correspondence points increases thanks to MKPC and the matching key-points in stage-2, making it easier for RANSAC to handle the noisy points generated by stage-1.
	
	\begin{algorithm}
		\caption{The workflow of the two-stage pipeline for image matching}
		\label{alg2}
		\begin{algorithmic}
			\REQUIRE ~~\\
			Two images to be matched $I_1$ and $I_2$; \\
			A set of models for stage-one $M^1=\{m^1_1, m^1_2, \cdots, m^1_{n_1} \}$; \\
			A set of models for stage-two $M^2=\{m^2_1, m^2_2, \cdots, m^2_{n_2} \}$; \\
			\ENSURE ~~\\
			(1) Define two empty sets of points $X^1_1$ and $X^1_2$ for both images in stage-one;\\
				\quad \ \  for ($m^1_i$ in $M^1$ ; $0<i<n_1+1$) \ do\\
				\quad \ \  \quad \ \  Input $I_1$ and $I_2$ into $m^1_i$; \\ 
				\quad \ \  \quad \ \  Get the matching key-points of $I_1$ and $I_2$ as $p^1_1$ and $p^1_2$; \\ 
				\quad \ \  \quad \ \ Extend $p^i_1$ into $X^1_1$, extend $p^i_2$ into $X^1_2$; \\
			(2) Input $I_1$, $X^1_1$,  $I_2$, and $X^1_2$ into MKPC algorithm and get the cropped critical regions $I'_1$ and $I'_2$; \\
			(3) Define two empty sets of points $X^2_1$ and $X^2_2$ for both images in stage-one;\\
			\quad \ \  for ($m^2_i$ in $M^2$ ; $0<i<n_2+1$) \ do\\
			\quad \ \  \quad \ \  Feed $I'_1$ and $I'_2$ into $m^2_i$; \\ 
			\quad \ \  \quad \ \  Get the matching key-points of $I'_1$ and $I'_2$ as $p'^1_1$ and $p'^1_2$; \\ 
			\quad \ \  \quad \ \ Extend $p'^i_1$ into $X^2_1$, extend $p'^i_2$ into $X^2_2$; \\
			(4) Scale the coordinates of $X^2_1$ into the scale of $I_1$; \\
			\quad \ \	Scale the coordinates of $X^2_2$ into the scale of $I_2$; \\
			(5) Concatenate $X^1_1$ and $X^2_1$ to get $X_1$; \\
			\quad \ \ Concatenate $X^1_2$ and $X^2_2$ to get $X_2$; \\
			(6) Calculate the F matrix / E matrix / H matrix by feeding $X_1$ and $X_2$ into RANSAC; \\
			(7) Return F matrix / E matrix / H matrix
			
		\end{algorithmic}
	\end{algorithm}

	\subsection{Crop Area Expansion (CAE)}

\begin{figure*}[!h]
	\centering
	%\fbox{\rule[-.5cm]{0cm}{4cm}  \rule[-.5cm]{4cm}{0cm}}
	\includegraphics[scale=0.2]{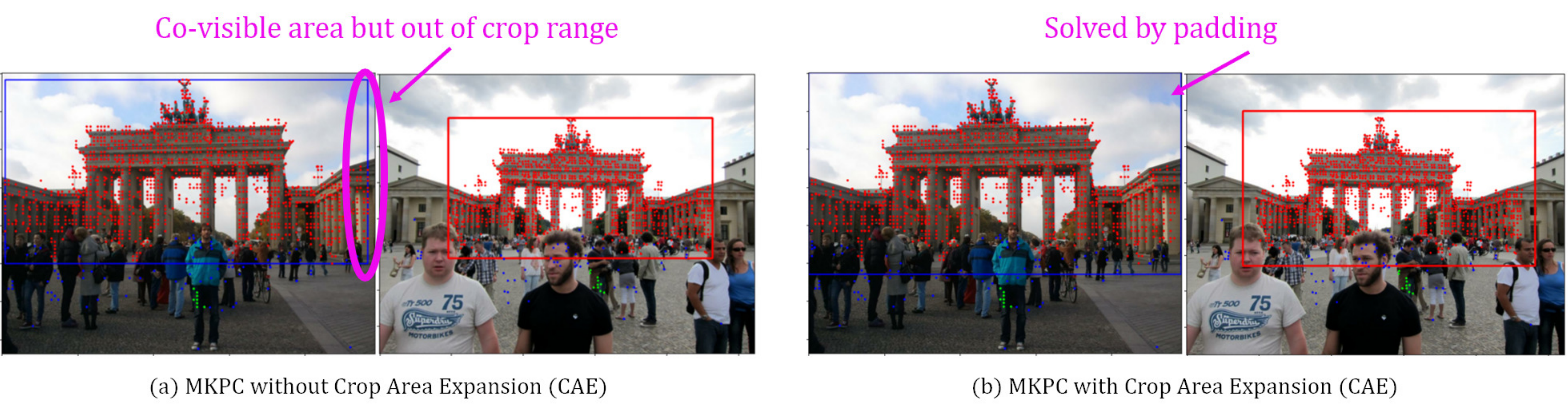}
	\caption{The effect of Crop Area Expansion (CAE). The CAE saves the area abandoned by MKPC incorrectly.}
	\label{pad}
	\end{figure*}

	Due to the structural problem of image matching models, the matching points may not always occur at the edges of objects. Therefore, cutting out only the outer perimeter of the cluster may miss some co-visible areas, wasting information as shown in Figure \ref{pad}(a). To this end, we designed a padding operations called Crop Area Expansion (CAE) to expand the critical regions obtained by MKPC. The CAE is controlled by two parameter, which is the horizontal expand factor $e_h$ and the vertical expand factor $e_f$. In Figure \ref{pad}(b), we set $e_h=1.05$ and $e_v=1.0$ for Image Matching Challenge 2022 benchmark, as there are usually informative regions in horizontal direction while the areas in vertical direction(buildings) are frequently information-poor(sky, ground, etc.). Thus, the CAE saves the area abandoned by MKPC incorrectly.

	\section{Experiments}
	
\begin{figure*}[!h]
	\centering
	%\fbox{\rule[-.5cm]{0cm}{4cm}  \rule[-.5cm]{4cm}{0cm}}
	\includegraphics[scale=0.14]{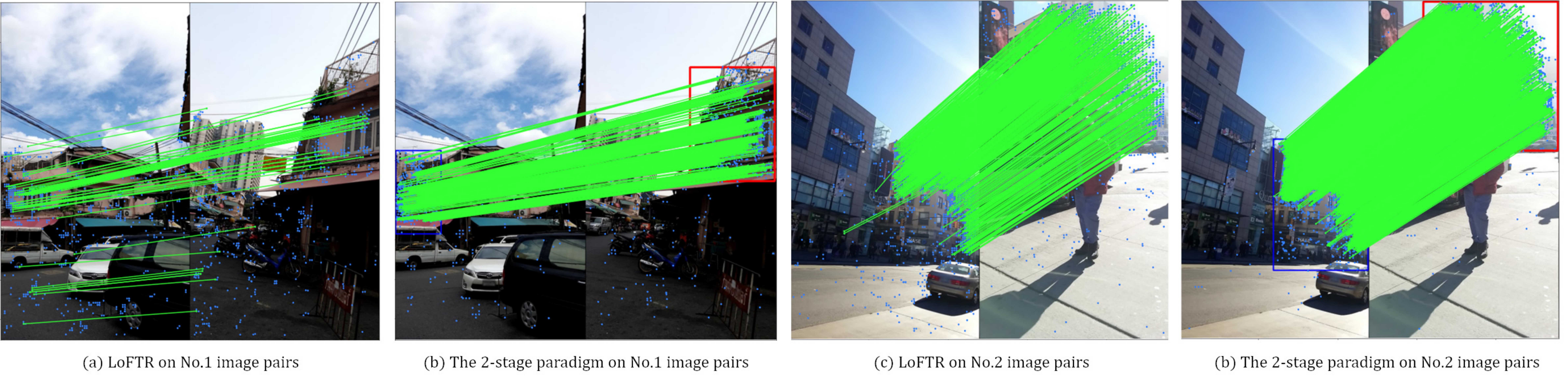}
	\caption{The Effect comparison of LoFTR and two-stage pipeline on two demo images pairs. The two-stage pipeline greatly reject the outliers, constrain regions of matching and refine the matching in critical areas.}
	\label{demo1}
	\end{figure*}

	As shown in Figure \ref{demo1}, The two-stage pipeline greatly reject the outliers, constrain area of matching and refine the matching in critical regions. We conduct experiment by plugging SuperPoint + SuperGlue into our two-stage pipeline on YFCC100M and ScanNet for outdoor and indoor pose estimation, respectively. What's more, we try plugging combinations of models into the two-stage pipeline, without any training, our method outperforms the SOTA of IMC2022 benchmark. 
	
	\subsection{MKPC with single model for outdoor scenario}
	
	\paragraph{Dataset and Metrics.} For outdoor pose estimation, we follow the experiments setting in SuperGlue~\cite{sarlin2020superglue}, evaluation on the PhotoTourism dataset, which
	is a subset of YFCC100M dataset~\cite{thomee2016yfcc100m} with ground truth poses. For the metrics, as ~\cite{brachmann2019neural, yi2018learning, zhang2019learning}, We take the AUC of the pose error at the thresholds ($5^{\circ}$, $10^{\circ}$, $20^{\circ}$) where the maximum of the angular errors evaluates the pose error
	in rotation and translation. The experimental results are recorded in Table \ref{spsg_yfcc}.
	where we try RANSAC and USAC\_MAGSAC (supported by OpenCV) to get the E matrix. As shown in Table \ref{spsg_yfcc}, the two-stage pipeline refine the performance of Superpoint + SuperGlue under the same sets of resolutions and the same parameters estimation algorithm by small but stable margin. Please note that we simply uses the pretrained weights on MegaDepth~\cite{li2018megadepth} for SuperPoint and SuperGlue. More specifically, under the same sets of resolutions and inference times, the two-stage pipeline is always better than ensemble directly. Without the constraint of keeping the computational complexity no larger than baselines, our approach brings a better enhancement.
	
	\begin{table}
		\small
		\caption{Use SuperPoint + SuperGlue with MKPC on YFCC. "Res" means resolution, which indicates the longest dimension after being resized. AUC@$5^{\circ}$ means the AUC of the pose error at the thresholds$=5^{\circ}$. "Params" means parameters.}
		\label{spsg_yfcc}
		\centering
		\begin{tabular}{cccccc}
			\toprule
			Res for stage-1 & Res for stage-2 &
			 Params estimation & AUC@5 & AUC@10 &AUC@20 \\
			\midrule
			1600 & None & RANSAC & 37.73 & 58.03 & 74.57  \\  
			%1600 & 1600 & RANSAC & 39.92 & 59.81 & 75.72 \\
			1600 & None & USAC\_MAGSAC & 38.29 & 58.33 & 74.58  \\
			1200, 1600 & None & USAC\_MAGSAC & 40.64 & 60.67 & 76.37 \\  
			1200 & 1600 & USAC\_MAGSAC & 40.88 & 60.58 & 76.26 \\
			1600 & 1200 & USAC\_MAGSAC & 40.94 & 60.78 & 76.26 \\
			%1600 & 1600 & USAC\_MAGSAC & 39.83 & 59.88 & 75.79 \\
			%1600 & 2000 & USAC\_MAGSAC & 38.3 & 58.78 & 75.29 \\
			%1600 & 1200, 1600 & USAC\_MAGSAC & 40.8 & 60.43 & 76.1 \\
			%1600 & 1600, 2000 & USAC\_MAGSAC & 40.06 & 60.22 & 76.04 \\
			%1600 & 1200, 1400, 1600 & USAC\_MAGSAC & 41.91 & 61.71 & 77.11 \\
			\midrule
			
			1200, 1400, 1600 & None & USAC\_MAGSAC & 41.51  & 61.22 & 76.73 \\
			1200, 1400 & 1600 & USAC\_MAGSAC & 41.66  & 61.48 & 76.72 \\
			1200, 1600 & 1400 & USAC\_MAGSAC & 41.95  & 61.52 & 76.80 \\
			%1200, 1600 & 1200, 1400, 1600 & USAC\_MAGSAC & 42.03  & 61.55 & 76.92 \\
			1600 & 1200, 1400 & USAC\_MAGSAC & 42.20  & 61.65 & 76.95 \\
			\midrule
			
			840, 1200, 1600 & None & USAC\_MAGSAC & 41.25 & 60.93 & 76.40 \\
			840, 1200, 1400, 1600 & None & USAC\_MAGSAC & 41.30 & 61.13 & 76.65 \\
			1600 & 840, 1200, 1400 & USAC\_MAGSAC & 42.10 & 61.42 & 76.79 \\
			840, 1200 & 1400, 1600 & USAC\_MAGSAC & 42.08 & 61.60 & 76.92 \\
			1200, 1600 & 840, 1400 & USAC\_MAGSAC & 42.23 & 61.92 & 77.20 \\
			840, 1200, 1600 & 1400 & USAC\_MAGSAC & 42.30 & 62.08 & 77.17 \\
			\midrule
			\textbf{840, 1200, 1600} & \textbf{1200, 1400, 1600} & \textbf{USAC\_MAGSAC} & \textbf{43.18} & \textbf{62.59} & \textbf{77.59} \\
			\bottomrule
		\end{tabular}
	\end{table}
	
	\subsection{MKPC with multiple models for outdoor pose estimation}
	Image Matching Challenge 2022 benchmark is an outdoor dataset, containing images pairs with variable degrees, which is often larger than those in other outdoor dataset. Each pair in IMC2022 benchmark is collected never less than a 24-hours interval, which is usually months or years apart, causing a domain gap. Bridging this domain gap is the main challenge in this benchmark. It requires researchers and competitors getting the fundamental matrix and uses mean Average Accuracy (mAA), which computes the error in terms of rotation(degrees) and translation(meters) for evaluation. As shown in Table \ref{imc}, our method outperforms the previous SOTA, which currently ranks the first place on both private leaderboard and public leaderboard of the benchmark. Please note that this challenge is held on Kaggle, which constraints the submission time within 9 hours. The winners of IMC 2022 all ensembles with multiple models, making their submission time close to the time limit, which means that the comparison between our method and theirs are fair. As shown in Table \ref{imc}, the Top-3 in IMC benchmark have significant advantages to other winners. However, our method outperforms the SOTA by 0.289\% on private leaderboard and 0.312\% on public leaderboard. On private leaderboard, the margin between our methods and \#rank1 is larger than that between \#rank1 and \#rank2. On public leaderboard, the margin is larger than that of \#rank1 and \#rank3. What's more, the margin between ours and \#rank4 is about two times to that between \#rank4 and \#rank11. 
	
	Our method brings more gain on IMC2022 Benchmark than SuperPoint + SuperGlue on PhotoTourism dataset because of three reasons. (i) The larger degrees with greater variance between images in IMC2022 maximizing the potential of our method, because as the degree increases, the necessity for extracting the co-visible area improves. (ii) The domain gap cause by the various collection time and environments constrain single model from fully capturing the common features between two images, making it's vital to ensemble multiple model for a better score on IMC 2022. Meanwhile, our approach is suitable for model ensemble naturally (iii) The comparison in Table \ref{imc} includes multiple model. As illustrated in the last reason, our model is significantly stronger when applied to ensemble. Appropriate ensemble in the first stage improves the quality of clipping in the ROI region, leading to better rejection of outliers and an increase in the upper limit in the stage-2. In stage-2, proper ensemble enhances the effectiveness of the models for fine matching within critical regions, ultimately producing better metrics.
	
	Finally, Both of the comparisons (the comparison between our method and other solutions on IMC2022 benchmark and the comparison between our method on IMC2022 benchmark and PhotoTourism) show not only the great effectiveness but also the strong generalization and robustness of our method.

	\begin{table}
		\caption{Our method on Image Matching Challenge 2022 Benchmark. We submit our method after the competition ends, the baselines are the winners solution on the final leaderboard in IMC 2022. The \#Rank1 represents the champion of IMC 2022, which is the SOTA on this benchmark. The \#Rank 1 to 11 were awarded gold medals by Kaggle, which are chosen as baselines. The public leaderboard takes 49 \% of the total IMC benchmark, while the private leaderboard takes 51 \%, while the private one is more an important, for which represents the ability of generalization of methods. The metric is mean Average Accuracy(mAA)}
		\label{imc}
		\centering
		\begin{tabular}{ccc}
			\toprule
			Methods  & mAA on Public Leaderboard & mAA on Private Leaderboard \\
			\midrule
			SuperPoint+SuperGlue & 0.67627 &0.67054 \\
			DKM & 0.66731 & 0.68651 \\
			LoFTR & 0.75478 & 0.75606 \\
			QuadTree & 0.81309 & 0.81223 \\
			AspanFormer & 0.79437 & 0.81438 \\
			\midrule
			\#Rank11 & 0.84783 & 0.84562 \\
			\#Rank10 & 0.83833 & 0.84611 \\
			\#Rank9 & 0.84803 & 0.84786 \\
			\#Rank8 & 0.84675 & 0.84871 \\
			\#Rank7 & 0.84287 & 0.84884 \\
			\#Rank6 & 0.85270 & 0.85023 \\
			\#Rank5 & 0.84824 & 0.85092 \\
			\#Rank4 & 0.85238 & 0.85225 \\
			\#Rank3 & 0.86100 & 0.85866 \\
			\#Rank2 & 0.86000 & 0.86210 \\
			\#Rank1 (previous SOTA) & 0.86367 &  0.86343 \\
			\textbf{Ours} & \textbf{0.86679} &  \textbf{0.86634} \\
			\bottomrule
		\end{tabular}
	\end{table}	

	\subsection{MKPC with single model for indoor pose estimation}

	ScanNet is a dataset for indoor pose estimation. Our experiments follow the set of SuperGlue~\cite{sarlin2020superglue}, whose results are in Table \ref{spsg_scan}. Under the same sets of resolutions and inference time, our approach can hardly surpass the naive ensemble. We think it's caused by the difference between the various morphology and texture distribution in indoor scenario (ScanNet) and outdoor scenario (YFCC). The indoor scenes always contain object with less texture, which is hard for image matching model. The effect of our two-stage pipeline strongly relies on the quality of the key-points generated by the modes in stage-one. Without key-points of great quality, MKPC is unable to propose the critical region with high accuracy. Moreover, the low scores of SuperPoint + SuperGlue on ScanNet reflects that only using SuperPoint + SuperGlue for stage-one may output key-points with lower quality, resulting in bad ROIs produced by MKPC. A poor ROI may instead have a negative impact on the result, rather than enhancement.
	
	\begin{table}
		\caption{Use SuperPoint + SuperGlue with MKPC on ScanNet. "Res" means resolution, which indicates the longest dimension after being resized. AUC@k means the AUC of top-k.}
		\label{spsg_scan}
		\centering
		\begin{tabular}{ccccc}
			\toprule
			Res for stage-1 & Res for stage-2 & AUC@5 & AUC@10 &AUC@20 \\
			\midrule
			640 & None & 16.18 & 34.11 & 52.47  \\  
			%640 & None & USAC\_MAGSAC & 16.02 & 33.63 & 52.46  \\  
			840 & None & 16.56 & 34.8 & 53.44  \\  
			640, 840 & None & 18.02 & 37.18 & 55.98  \\  
			640, 840, 1024 & None & 19.59 & 38.37 & 57.47  \\  
			640, 840, 1024, 1280 & None & 19.84 & 38.93 & 57.00  \\  
			640, 840, 1024, 1280, 1536 & None & 20.33 & 39.04 & 57.15  \\  
			%840 & 640 & RANSAC & 17.45 & 35.69 & 53.79  \\  
			840, 1024, 1280 & 840, 1280 & 20.18 & 38.76 & 56.94  \\  
			%640 & 840 & RANSAC & 17.62 & 36.88 & 55.79 \\
			%640 & 1024 & RANSAC & 17.85 & 35.86 & 55.05 \\
			%\textbf{640} & \textbf{640, 840, 1024} & \textbf{RANSAC} & \textbf{17.94} & \textbf{36.87} & \textbf{55.56} \\
			%\textbf{840, 1024, 1280} & \textbf{840, 1024, 1280, 1536} & \textbf{RANSAC} & \textbf{19.56} & \textbf{37.94} & \textbf{55.97} \\
			\bottomrule
		\end{tabular}
	\end{table}	
	
	\subsection{The Time-consumption of the MKPC and the Two-stage Pipeline}
	Although our method requires models to infer for multiple times, the time-consumption is not as large as it looks like for the following reasons. (i) The MKPC is a plug-and-play module and the two-stage pipeline is campatible to any image matching models, while users don't require to re-train or fine-tune any image matching models. This advantage makes it a convenient choice for improving the effect of image model without any training. (ii) The parameters estimations algorithm such as RANSAC filter outliers and find the global optimized F/E/H matrix within a maximum iterations, which will stop if the matching error decreases below the threshold in early stage. In stage-two, the critical regions are areas proposed by MKPC, making the key-points generated by the models in stage-two have a much higher quality. Hence, the quality of sets of key-points generated by our two-stage pipeline are usually better than those produced by single model or naive ensemble, making RANSAC may stop in an earlier stage (iii) The DBSCAN consumes little time and computational resource, which is almost negligible. Please note that the time-consumption various from datasets, which is essentially influenced by the distribution of the original images.

	\section{Conclusion}
	In this paper, we first introduce an hypothesis that the area outside the co-visible regions of both images contain little information. To narrowing the restrain space and reject outliers, we propose a plug-and-play but remarkable algorithm called matching key-points crop (MKPC). Based on MKPC, we further present a two-stage pipeline which is compatible to any image matching models and able to strengthen any image models without re-training or fine-tuning. The experiments indicate that our method has a small but promising effect for outdoor pose estimation, while it still works on indoor scenario but requires more computational complexity. What's more, our method surpassing the SOTA of Image Matching Challenge 2022 Benchmark, indicating the potential and benefits of our approach when implemented for ensemble, and the effectiveness and generalization of our method when applying for multiple models.
	
	\section{Future Work}
	In future work, we aim to address the limitations and challenges observed in our experiments on various datasets. Specifically, we plan to (i) explore the potential of our method in model fusion by incorporating multiple image matching models with our approach, particularly on benchmarks like IMC2022 where image pairs with larger angle deviations and domain gaps, as our approach seems particularly suitable for these scenarios. (ii) Extending our two-stage pipeline to newer and more advanced models beyond SuperPoint+SuperGlue, as the accuracy of the first stage is crucial for our method. We expect that conducting further experiments on PhotoTourism(subset of YFCC100M) and ScanNet datasets using stronger baselines or ensemble of models will help us better understand our method and enhance the performance across various scenarios.
	
	\newpage
	
	\bibliographystyle{IEEEbib}
	\bibliography{mkpc}
	
\end{document}